 \documentclass[final,5p,times,twocolumn]{elsarticle}

\usepackage{amssymb}
\usepackage{leftidx}
\usepackage{booktabs}
\usepackage{setspace}

\usepackage{cases}
\usepackage{color}

\usepackage{lineno}
\usepackage[perpage,symbol]{footmisc}

\begin{document}

\begin{frontmatter}



\title{Towards Direct Medical Image Analysis without Segmentation}

\author{Xiantong Zhen$^{a,b}$ and Shuo Li$^{a,b,c}$\corref{cor}}
\cortext[cor]{Corresponding author (slishuo@gmail.com).}
\address{$^a$The University of Western Ontario, London, ON, Canada.\\
$^b$The Digital Image Group (DIG), London, ON, Canada.\\
$^c$Lawson Health Research Institute, London, ON, Canada.\\
}

\end{frontmatter}
Direct methods have recently emerged as an effective and efficient tool in automated medical image analysis and become a trend to solve diverse challenging tasks in clinical practise. Compared to traditional methods, direct methods are of much more clinical significance by straightly targeting to the final clinical goal rather than relying on any intermediate steps. These intermediate steps, e.g., segmentation, registration and tracking, are actually not necessary and only limited to very constrained tasks far from being used in practical clinical applications; moreover they are computationally expensive and time-consuming, which causes a high waste of research resources. The advantages of direct methods stem from \textbf{1)} removal of intermediate steps, e.g., segmentation, tracking and registration; \textbf{2)} avoidance of user inputs and initialization; \textbf{3)} reformulation of conventional challenging problems, e.g., inversion problem, with efficient solutions.

Direct methods in medical image analysis scenarios are defined as a series of methodologies that estimate clinical measurements directly from medical imaging data without relying on any unnecessary, intermediate steps. The principle behind direct methods is that there are intrinsic relationship existing between medical images and clinical measurements; these relationship can be directly extracted and modeled to estimate clinical measurements for diagnosis and prognosis without relying on any intermediate stages.

Direct methods have recently demonstrated its great effectiveness and efficiency in many aspects for direct medical image analysis, especially on two important applications: volume estimation and functional analysis.

\textbf{Direct estimation} of cardiac volumes without segmentation has shown remarkable effectiveness due to its great advantages over conventional segmentation based methods. For more than 20 years, cardiac volume estimation has long been suffering from the intermediate, unreliable and even intractable segmentation steps in traditional methods. Segmentation has only been focused on a single ventricle, e.g., the left ventricle (LV) on a bi-ventricular view for a long time, and recently started to work on the right ventricle (RV), which remains unsolved, not to mention joint bi-ventricles, i.e., LV and RV, and even more challenging four chambers, i.e., LV, RV, LA and RA. All the four ventricular volumes, however, are routinely and intensively used in clinical practise for cardiac functional analysis.

Direct volume estimation removes segmentation and estimate the volume directly from images by capturing and disentangling the hihgly non-linear relationship between image appearance and volumes. Direct estimation provides a general framework that can flexibly deal with a single ventricle \cite{afshin2012global}, joint bi-ventricles \cite{wang2013direct, zhen2014direct, zhen2015multi} and simultaneous four chambers \cite{zhen2015four} in a more convenient way within a unified single framework.

\textbf{1)} The initial success of direct volume estimation has been demonstrated by LV volume estimation in \cite{afshin2012global}. Image features by statistics based on the Bhattacharyya coefficient of similarity between image distributions; it is demonstrated that these statistics are non-linearly related to the LV volumes and can be used for ejection fraction estimation via neural networks directly.

\textbf{2)} Futher success of direct volume estimation has been achieved for joint bi-ventricular volume estimation \cite{wang2013direct, zhen2014direct, zhen2015multi}. The first attempt \cite{wang2013direct} adopts a Bayesian model in which the LV and RV volumes are calculated as the weighted average over the templates. Random forests are introduced in \cite{zhen2014direct} for more efficient joint bi-ventricular volume estimation without relying on assumed models. Recently, a fully data-driven, learning-based framework for bi-ventricular volume estimation has been developed in \cite{zhen2015multi}.

\textbf{3)} Direct volume estimation has been generalized and validated on multiple ventricles for cardiac four chambers in \cite{zhen2015four}. Based on supervised descriptor learning, multi-output regression forests are adopted to directly and simultaneously estimate four chamber volumes. This achieves for the first time achieves general framework for direct volume estimation of either a single ventricle or multiple ventricles.


\textbf{Direct diagnosis} has started to generate increasing interest \cite{wang2013direct, zhen2014direct} in cardiac image analysis which has long been severely obstructed by segmentation, registration and tracking in conventional approaches.

\textbf{1)} Direct diagnosis of cardiac functional abnormalities has further demonstrated the power of direct methods \cite{afshin2014regional, zettinig2014data}. Traditional methods are highly dependent on segmentation, registration and tracking \cite{punithakumar2013regional}, which are conditionally expensive and inevitably induces cumulative errors. A direct method for regional assessment of the cardiac LV myocardial function via classification has been proposed in \cite{afshin2014regional}. The method formulates diagnosis as a classification problem based on statistical features which are related to the proportion of blood within each segment and can characterize segmental contraction. Without tracking or segmentation of regional boundaries, they obtain an direct assessment of cardiac segmental abnormality in real-time with more accurate results than segmentation-based methods.

\textbf{2)} Direct diagnosis has also been developed to resolve traditional challenging problems, e.g., inverse problems, with more efficient solutions. Diagnosis and treatment of dilated cardiomyopathy (DCM) is challenging due to the large variety of disease stages. Cardiac models of cardiac electrophysiology (EP) is commonly used to conduct the challenging personalization of model parameters. Traditional methods require to solve an inverse problem which is challenging and computationally demanding. Direct diagnosis and treatment of DCM has been explored in \cite{zettinig2014data} which achieves data-driven estimation of cardiac electrical diffusivity from 12-lead ECG. The direct method learns the inverse function by formulating as a polynomial regression problem. Compared to traditional approaches, the direct method can find model parameters directly from EEG signals for specific patients and provides more accurate results in a more efficient way.

In addition to the great success of direct methods in volume estimation and diagnosis, direct methods have also been developed in many other important medical image applications, e.g., anatomy detection/localization \cite{criminisi2013regression}, shape analysis \cite{zhou2010shape} and clinical pattern prediction \cite{namburete2015learning}. The great effectiveness and uprising role of direct methods in medical image analysis and clinical practise stem from multiple attractive merits in contrast to traditional indirect methods:

\vspace{-2mm}
\begin{itemize}
\item Direct methods enable to focus on the final clinical goal without relying on any unnecessary, unreliable, intermediate steps. By directly modeling the relationship between image appearance and medical measurements of clinical interest, direct methods estimate clinical measurements directly from medical images in a more convenient, efficient and accurate way. Therefore, direct methods can not only save high computational cost but also avoid errors induced by any intermediate operations, which enables practical use in clinical applications.

\vspace{-2mm}
\item Direct methods provide more efficient resolutions of conventional problems, e.g., inversion problems, by statistical learning. In contrary to their conventional counterparts, statistical models have more compact and exquisite formations and are highly generalized, and therefore possess extensive flexibility and generality for different clinical applications. Moreover, due to the statistical learning, direct methods also enable us to richly explore data statistics which therefore offers more meaningful and comprehensive clinical assessment.

\vspace{-2mm}
\item Direct methods serve as a bridge between rapidly-updating machine learning algorithms and medical image analysis, which accelerates the translation of fundamental research to clinical practise. Direct methods allow to leverage advanced machine learning techniques from artificial intelligence for medical image analysis. By deploying machine learning algorithms, direct methods are able to not only explore distinguishing image features that are closely related to diseases but also take advantages of papulation-based analysis that helps accurate diagnosis.
\end{itemize}

The established success of direct methods on diverse applications has validate its effectiveness in medical image analysis and shown the great potential in clinical practice. Due to the attractive merits of efficiency, effectiveness and convenience, direct methods provide innovative solutions for traditional medical image analysis and will promise its great perspective in even broader clinical applications, e.g., volume estimation, functional analysis, clinical prediction and diagnosis. With direct methods, traditional time-consuming tasks will be replaced with more efficient methods; those complex tasks will be simplified and handled in a more convenient way; and those intractable tasks will be resolved quickly with more effective direct formulations.

Direct methods will continue to show its legendary on solving even more challenging clinical tasks with the rapid development and advancement of machine leaning techniques in artificial intelligence, and proliferation of large amount of medical imaging data. In the foresee future, we can expect the marriage of direct methods with advanced state-of-the-art machine leaning techniques, e.g., deep learning and statistical discriminative learning, which will further boost the development of direct medical image analysis and promote its wide use in clinical practice.

\end{document}